\documentclass[letterpaper, 10 pt, conference]{ieeeconf}
\IEEEoverridecommandlockouts
\usepackage{amsmath,amssymb,amsfonts}
\usepackage{algorithmic}
\usepackage{graphicx}
\usepackage[export]{adjustbox}
\usepackage{textcomp}
\usepackage{xcolor}
\usepackage{url}
\usepackage{array}
\usepackage{tabularx}
\usepackage{makecell}
\usepackage{booktabs}
\usepackage{multirow}
\usepackage{hyperref}
\usepackage[caption=false]{subfig}
\usepackage{tikz}
\usepackage{pgfplots}
\usepackage{threeparttable}
\DeclareMathOperator*{\avg}{avg}
\DeclareMathOperator{\pow}{pow}

\def\BibTeX{{\rm B\kern-.05em{\sc i\kern-.025em b}\kern-.08em
    T\kern-.1667em\lower.7ex\hbox{E}\kern-.125emX}}

\title{Developing Combined Manipulation and Locomotion Skills with Interaction Representation and Skill Composition} 

\author{Fanxing Meng$^{1}$ and Jing Xiao$^{1}$
\thanks{$^{1}$Robotics Engineering Department,
        Worcester Polytechnic Institute, Worcester, MA, USA
        {\tt\small \{fmeng, jxiao2\}@wpi.edu}}%
}

\begin{document}
\maketitle
\begin{abstract}
This paper addresses how to enable a humanoid robot to learn motion policies based on developmental principles and combine policies to create more sophisticated and useful behaviors. Specifically, we present an approach to (1) learning a whole-body reaching and grasping policy and (2) combining it and a standing-up and walking policy to compose a more complex policy of manipulation and locomotion: grasping, standing up, and walking.
In (1), our method draws inspiration from harmonic analysis and adopts cubic harmonics as weights to represent the hand-object spatial relationship via spatial convolution. Utilizing an intra-episode finger joint decoupling curriculum based on developmental principles, a robot can autonomously learn a generalizable grasping policy without relying on external datasets or pretrained models. 

In (2), our method combines the grasping policy with a separately learned getting-up policy by providing both policies with their respective observation vectors and using hand-object interaction scores to determine when each policy should control which robot joints. Our results show a 93\% zero-shot success rate for grasping unseen objects and a 96--100\% success rate for standing up while holding the object. Our work also demonstrates that combining different policies is only effective if each policy learning happens on the same whole humanoid body even if a policy (such as for locomotion) does not seem to need all the body parts (such as fingers). \footnote{Video available at \url{https://youtu.be/x-7x89fSJWY}}
\end{abstract}

\section{Introduction}
Human infants are capable of learning motor skills through self-exploration and intrinsic motivation. They can also combine different skills into unified behaviors such as crawling to a toy, picking it up, and carrying it around. Developmental science suggests that motor development proceeds in stages: rolling precedes crawling, crawling precedes standing, and reaching precedes grasping~\cite{Cangelosi2015Developmental}. Developmental robotics focuses on applying baby-inspired learning principles to robotics, with previous work enabling humanoid robots to autonomously learn to roll, kneel, stand, and walk in sequence by structuring the learning process as a graph~\cite{Meng202XFrom}. We aim to extend this approach to object reaching and grasping using a humanoid robot model with anthropomorphic hands.

Developing skills to interact with objects and blending them with other whole-body motor skills raises {\bf two new questions}. The first is representational: {\em what type of spatial information about robot-object interaction is efficient and effective for learning without large-scale human demonstration data or pretrained vision-language models}? The second is architectural: {\em once separate skills (e.g., reach, grasp, stand) are learned, how can they be combined to create a continuous, more complex whole-body behavior}?

To address the representation question, we hypothesize that a distance field can represent an object's shape; the spatial positions of the robot's fingers within the object's distance field represent the hand-object interaction relation. Distance field values alone provide only a radial component; the angular distribution of the fingers surrounding the object also needs to be captured. To encode the angular component, we draw inspiration from the shapes of atomic orbitals, which describe the probabilistic locations of electrons in atoms. We perform spatial convolution over the radial and angular components of all fingers around the object as a low-dimensional embedding, rather than using them disjointedly. Our experiments show that convoluted finger-object distances and orientations are more advantageous for grasp success rate and learning speed than unconvoluted raw values.

To address the architecture question, we run multiple policies simultaneously and use subtask progress to determine which policy controls which robot joints. This brings up a question regarding cross-embodiment transferability: can a stand-up policy trained on a robot model without fingers be directly applied to a robot model with fingers by only letting it control the body joints (but not the hands)? Our experiments produce a counter-intuitive outcome: the missing fingers caused an embodiment mismatch that even domain randomization could not resolve.

This research makes three major contributions:
\begin{itemize}
    \item Establishing a set of learning tools for reaching and grasping, including a reversible achievement-triggered reward graph and gradual finger joint decoupling, inspired by developmental principles.
    \item Formulating a cubic-harmonics-weighted spatial convolution with inversely scaled distance fields as a compact representation of the hand-object interaction relation.
    \item Composing manipulation and locomotion policies on a humanoid robot to perform object reaching and grasping, followed by standing up and walking, and verifying the necessity of training on the complete body for plug-and-play composition without extra adaptation.
\end{itemize}

\section{Related Work}

This work draws on three areas: developmental robotics, spatial representations for grasping, and reinforcement learning (RL) policy composition. We briefly describe them and indicate how this work differs from the prior literature.

\subsection{Developmental Robotics}
Piaget's theory suggests that learning is more effective through a curriculum, allowing the learner to start by mastering foundational skills that are prerequisites for more complex skills~\cite{Piaget}. Prior work implemented this idea in reinforcement learning using an achievement-triggered reward graph, where each node represents a motor skill (rolling over, kneeling, crawling, etc.) as a reward function; the edges of the graph represent achievement conditions between motor skills: after a parent node's reward value has reached a passing score, its child nodes' reward values can be incorporated into the total reward~\cite{Meng202XFrom}. Although this formulation works well for gross motor skills where different skills require different poses, it lacks the ability to define an enforcement mechanism to require a prerequisite skill like pre-grasping to sustain after having reached a passing score.

Developmental robotics methodologies also tried to simplify high-degree-of-freedom (DoF) learning problems by using a curriculum to unlock distal joints after proximal joints~\cite{proximal}, learning gross motor skills before fine motor skills. However, prior work only applied this methodology to simple grippers~\cite{gripper} or unlocked all hand joints at once~\cite{handjoints}.

Our work aims to extend these methodologies to whole-body reaching and grasping by incorporating an enforcement condition into the achievement-triggered reward graph and a curriculum that gradually unlocks the finger joints.

\subsection{Spatial Representations for Grasping}
Recent work in manipulation learning has leveraged distance fields in favor of point clouds or voxels as an intermediate representation of the object scene due to their rich geometric structure~\cite{sdf}. However, the complete distance field is difficult to use directly by a reinforcement learning policy. To address this issue while allowing generalization across different object types, existing work often encodes the distance field as a neural network~\cite{isdf} or learns neural grasping distance functions from pose databases~\cite{ngdf}. However, these approaches require training with large datasets to generalize.

In other domains such as biochemistry and crystallography, harmonic decomposition has often been applied to spatial mapping and shape representation. Spherical harmonics, for example, are used to reconstruct anatomical boundaries from discrete tomographic data~\cite{tomography}. Cubic harmonics, which are real-valued linear combinations of spherical harmonics, are used in modeling multipolar electronic populations~\cite{multipoles}. In robotics, however, harmonic decomposition has only been applied in path planning~\cite{pathplanning}, navigation~\cite{navigation}, and system control using isotypic subspaces~\cite{isotypic}.

Our work attempts to combine the strengths of distance fields and harmonic functions via spatial convolution to analytically compute a generic grasp interaction representation that does not require pre-training.

\subsection{RL Policy Composition}
RL policies for manipulation and locomotion are often trained separately due to their different goals and reward definitions. Existing work often relies on hierarchical reinforcement learning, which uses a high-level policy to output subgoals and weights for low-level policies~\cite{skillblender}. A simpler idea is multi-expert synthesis, which provides each expert policy with a minimal state observation vector optimized for the specific task~\cite{mes}. However, these approaches require training an additional meta-policy or high-level controller. 

Our policy composition method instead uses a quantitative transition threshold to determine when each policy should take control of which joint, offering flexibility without requiring an additional trained meta-policy.

\section{Method}
This section provides an overview of the problem formulation and introduces each component of our approach.

\subsection{Problem Formulation}
We use reinforcement learning to train a neural network policy on a humanoid robot model to perform whole-body object reaching and grasping before standing. The policy takes an observation vector comprising environment states such as limb poses and joint forces as input and outputs an action vector in terms of joint torque commands. The reward functions are evaluated to serve as the learning signal that the learner tries to maximize. At inference time, the trained grasping policy will be composed with a getting-up policy to stand up and walk while keeping the object grasped.

Our approach focuses on acquiring the spatial reasoning ability for grasping; hence, we bypass the computer vision step and assume that the objects models are obtained via multi-view perspective and provided to a 3D preprocessing library to compute their distance fields.

To provide a generic methodology, we do not presume any specific existing RL algorithm. Our methods only involve pre-computing the observation vector, pre-processing action outputs before applying them to robot joints, and designing the reward functions, which makes it agnostic to the RL agent implementation or neural network architecture.
 
\subsection{Inversely Scaled Distance Field (ISDF)}
In robotics and computer vision, signed distance fields are often used to represent the shortest distance of an arbitrary point $\mathbf{i}$ in space to the surface of an object $\partial\Omega$ as a continuous function $S: \mathbb{R}^3\rightarrow\mathbb{R}$. The value of $S(\mathbf{i})$ is positive when $\mathbf{i}$ is outside the object, negative if $\mathbf{i}$ is inside the object, and 0 if $\mathbf{i}$ is located on the boundary.

To model proximity as a reward, we define an inversely scaled distance field by applying an inverse exponential transformation to the signed distance field, producing a smooth attractive potential around the object:
\begin{equation} \label{eq:1}
I(\mathbf{i}) = \min{\left(1, \exp{\left(-\alpha S(\mathbf{i})\right)}\right)}
\end{equation}
where $\alpha$ is a fixed positive hyper-parameter controlling the curvature of the transformation. The function ignores negative distances (penetration) by capping the result at 1.

\subsection{Cubic Harmonics Weighted Spatial Convolution (CHWSC)}
\begin{figure}[htbp]
\vspace{-15pt}
    \centering
\subfloat[$\ell=4$, $m=-1$]{\includegraphics[clip, trim=40px 40px 800px 40px, width=0.16\textwidth]{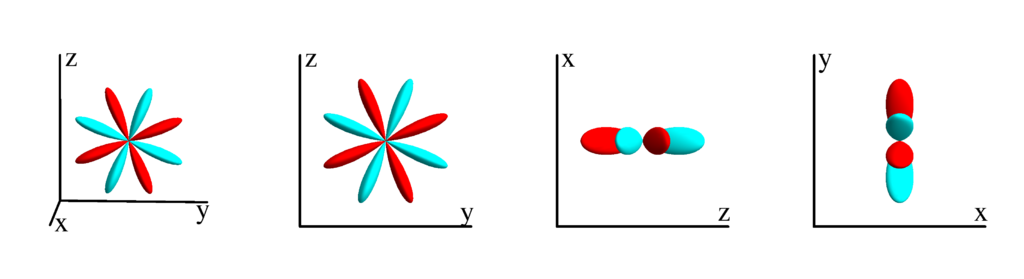}}\hfill
\subfloat[$\ell=5$, $m=4$]{\includegraphics[clip, trim=40px 40px 800px 40px, width=0.16\textwidth]{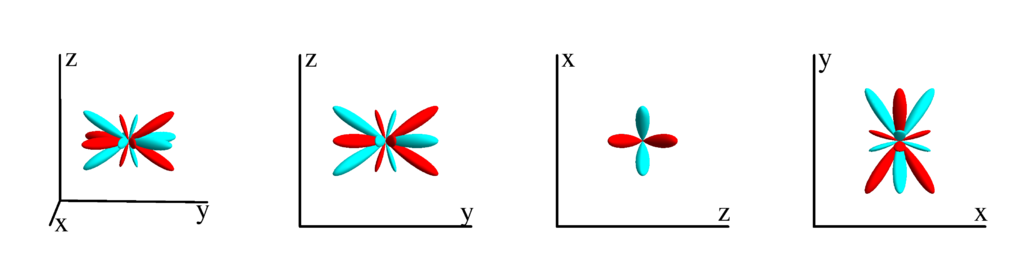}}\hfill
\subfloat[$\ell=6$, $m=-5$]{\includegraphics[clip, trim=40px 40px 800px 40px, width=0.16\textwidth]{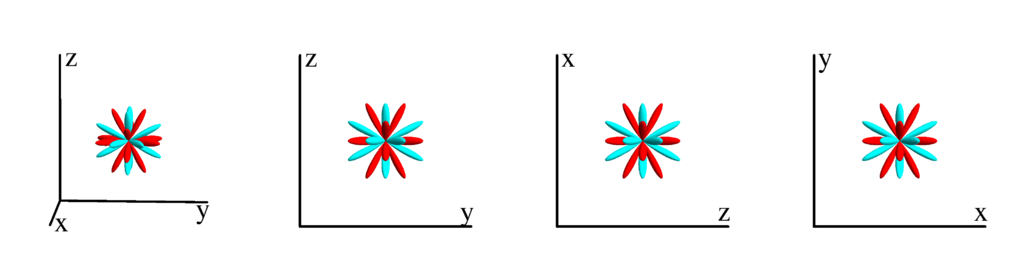}}

\caption{Surface representations of selected cubic harmonics. The phase is indicated by color, with \textbf{red} representing positive values and \textbf{cyan} representing negative values.\protect\footnotemark}
\label{cubic}
\end{figure}

Harmonic functions are the solutions to Laplace's equation. In spherical coordinates, the solutions are spherical harmonics, a family of complex-valued functions denoted by $Y^m_\ell(\theta, \varphi)$, where $\ell$ is a non-negative integer representing the degree and $m$ is the order, which can take values between $-\ell$ and $\ell$. $\theta$ and $\varphi$ are spherical angular coordinates. For computational benefits, it is common to use cubic harmonics, which are real-valued linear combinations of spherical harmonics, denoted as $K^m_\ell(\mathbf{r})$, where $\mathbf{r}$ is the Cartesian coordinate vector corresponding to $(\theta, \varphi)$. Fig.~\ref{cubic} visualizes the angular distribution using the radius of the surface to represent the magnitude of the cubic harmonics in that direction.

We use cubic harmonics as weights\footnotemark to convolute with the inversely scaled distance from each finger segment $\mathbf{f}_j$ to the object, represented in the object's reference frame:
\begin{equation}
    \mathrm{CHWSC}(l) = \sum_{\ell=0}^l\sum_{j} I(\mathbf{f}_j) \, K_\ell^m\left(\frac{\mathbf{f}_j}{\left\|\mathbf{f}_j\right\|}\right)
\end{equation}
\footnotetext{Excluding asymmetry ones such as $K^{-1}_2$, $K^{-4}_4$, $K^{-2}_4$, $K^{-6}_6$, $K^{-3}_6$ due to out-of-range numerical values}
\footnotetext{Images sourced from \url{https://www.quanty.org/physics_chemistry/orbitals/k}}

\subsection{Hand-object Interaction Scores}
To quantitatively evaluate grasp quality and provide dense reward to the learning agent, we define the following five metrics, illustrated in Fig.~\ref{fig:interaction_scores}.

\subsubsection{Hand-object proximity}
We use $I(\mathbf{h})$
to represent how close the hand is to the object, where $I()$ is defined in Eq.~\ref{eq:1} and $\mathbf{h}$ is a fixed point on the hand, as shown in Fig.~\ref{fig:hand_obj_proximity}.

\subsubsection{Interior-normal alignment}
We use $-\vec{\mathbf{n}_\mathrm{f}}\cdot\vec{\mathbf{n}_\Omega}$
to encourage each finger segment's inward-pointing surface normal ($\vec{\mathbf{n}_\mathrm{f}}$) to be anti-parallel to the object surface normal ($\vec{\mathbf{n}_\Omega}$) closest to it, as shown in Fig.~\ref{fig:interior_normal}.

\subsubsection{Finger-object proximity}
We use $\avg_j(I(\mathbf{f}_j))$
to represent how close each finger segment is to the object, where $\mathbf{f}_j$ is a fixed point on a finger segment, as shown in Fig.~\ref{fig:finger_obj_proximity}.

\subsubsection{Object-finger proximity}
In contrast to finger-object proximity, which measures how close each finger is to any part of the object, this metric evaluates how close each sampled point on the object is to any finger, as shown in Fig.~\ref{fig:obj_finger_proximity}. We denote this as
$$\avg_k\left(\exp{\bigl(-\alpha\;\min_{j}\bigl\|\mathbf{f}_j-\mathbf{\Omega}_k\bigr\|\bigr)}\right)$$
where $\left\|\mathbf{f}_j-\mathbf{\Omega}_k\right\|$ is the distance from a point $k$ on the object to the nearest finger segment $j$. The points are sampled on the object surface by curvature matching: the density of the point distribution is highest where the surface curvature is closest to the characteristic length of the hand. 

\subsubsection{Encapsulation}
\begin{figure*}[htbp]
    \centering
    \subfloat[Hand-object proximity]{\includegraphics[width=0.19\linewidth]{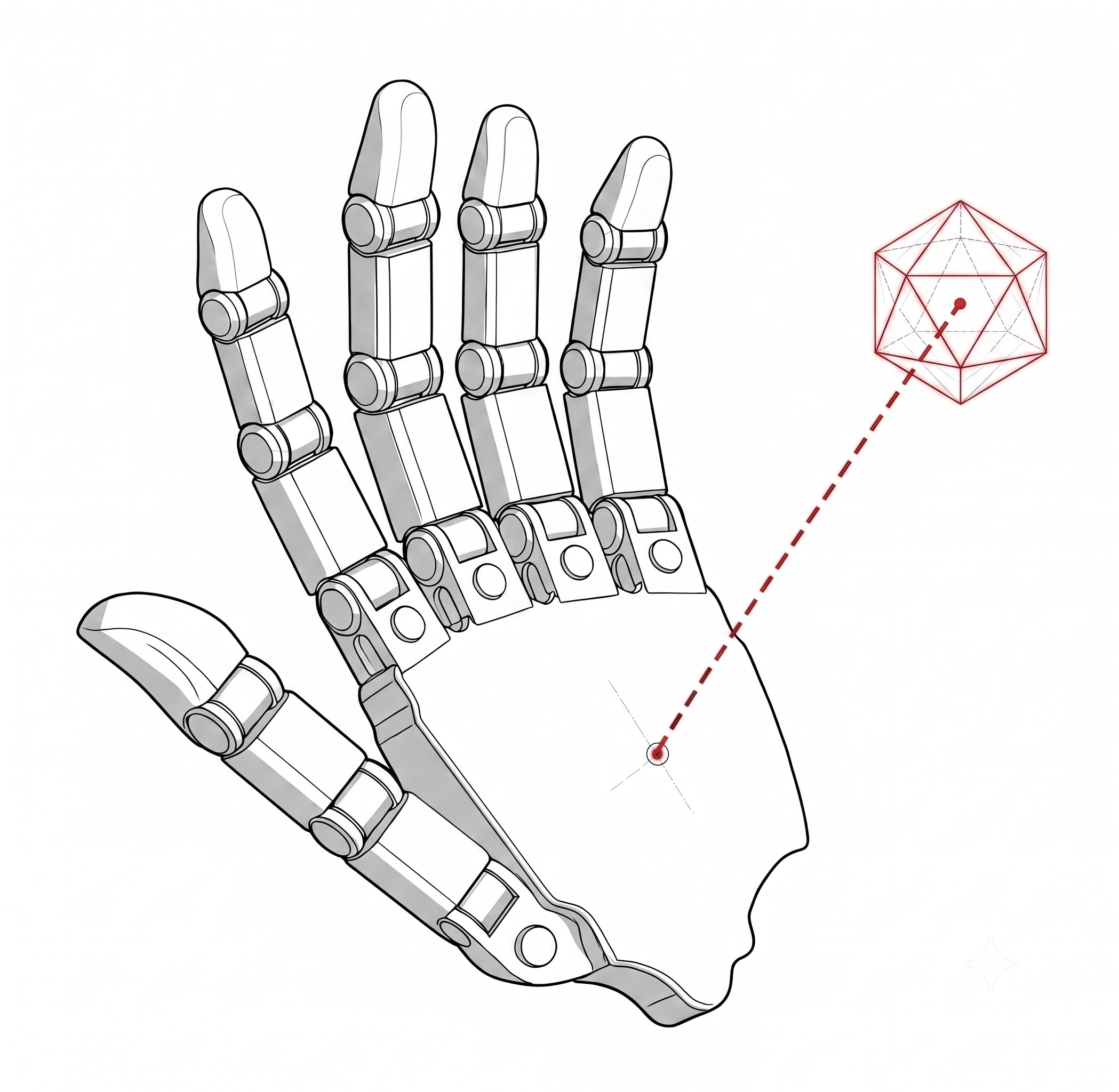}\label{fig:hand_obj_proximity}}
    \hfil
    \subfloat[Interior-normal\protect\\alignment]{\includegraphics[width=0.19\linewidth]{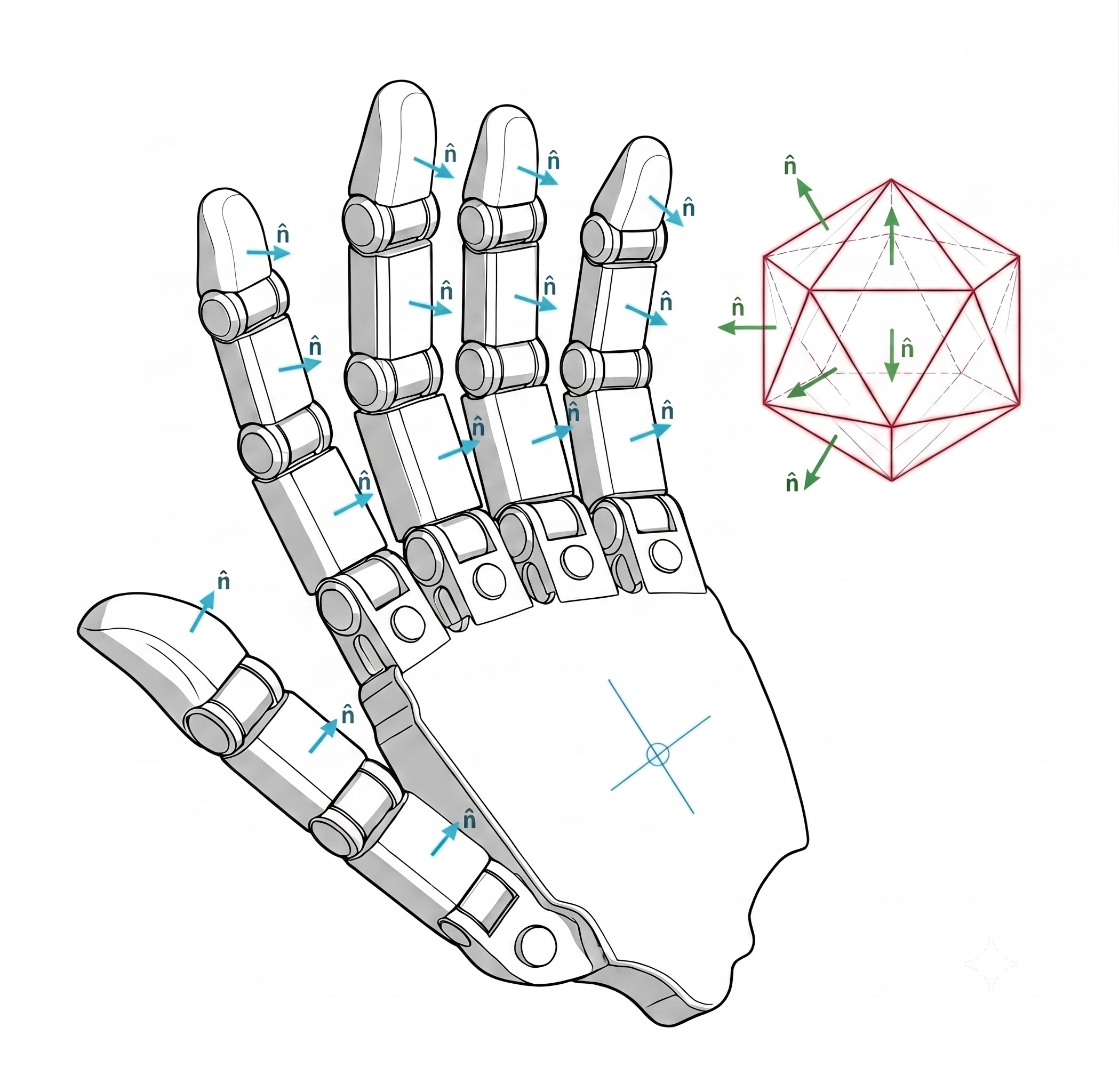}\label{fig:interior_normal}}
    \hfil
    \subfloat[Finger-object\protect\\proximity]{\includegraphics[width=0.19\linewidth]{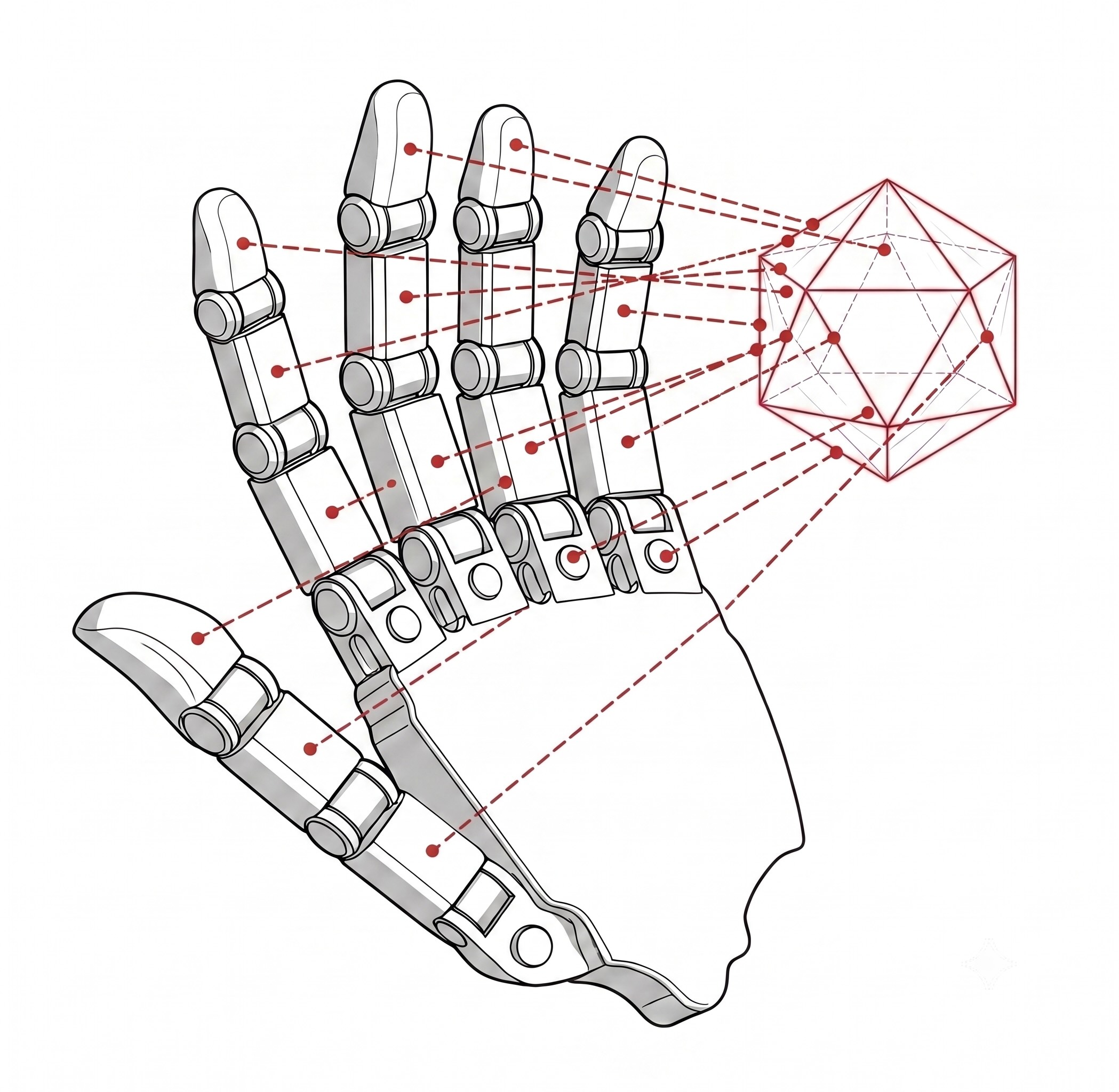}\label{fig:finger_obj_proximity}}
    \hfil
    \subfloat[Object-finger\protect\\proximity]{\includegraphics[width=0.19\linewidth]{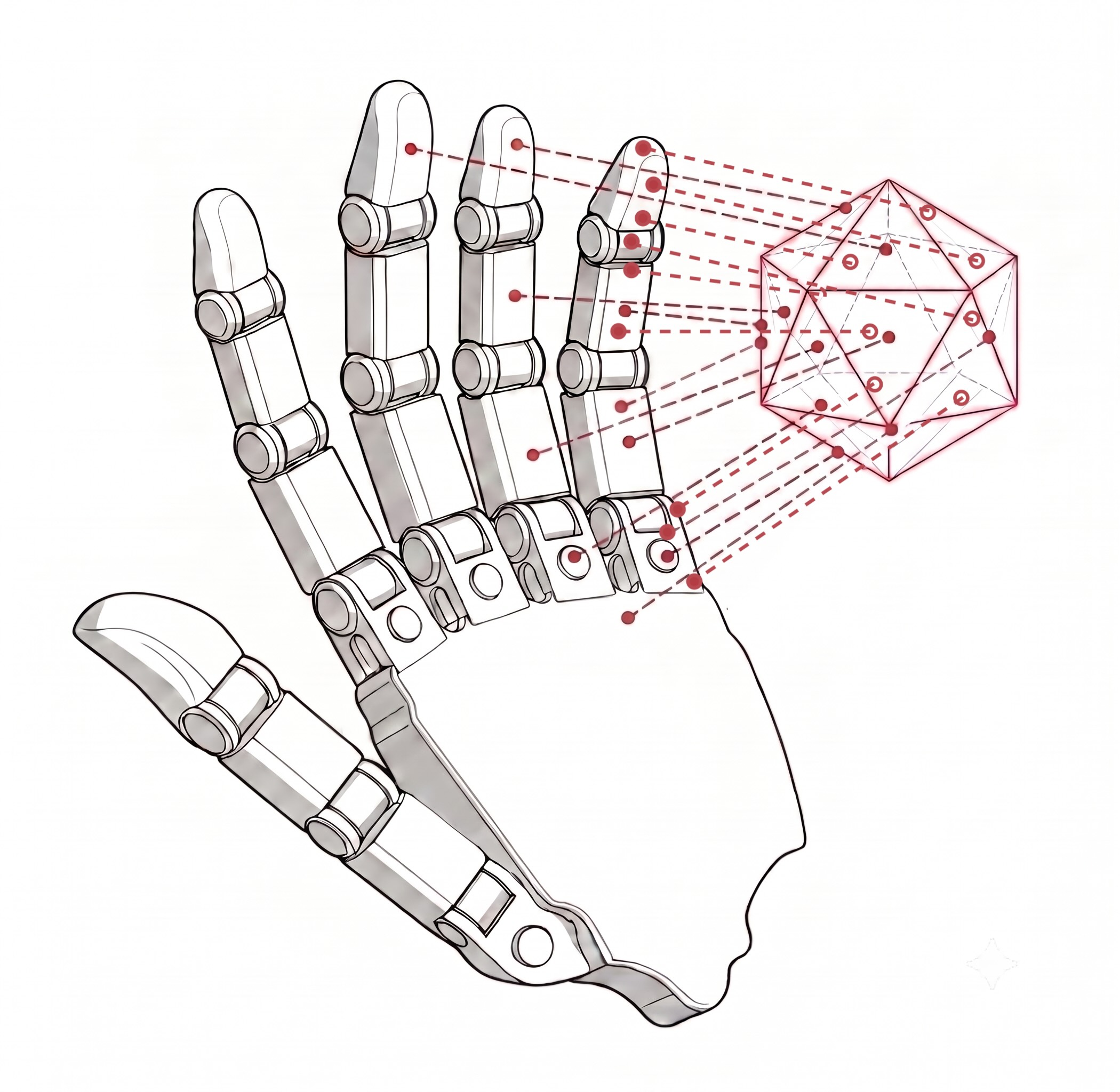}\label{fig:obj_finger_proximity}}
    \hfil
    \subfloat[Encapsulation]{\includegraphics[width=0.19\linewidth]{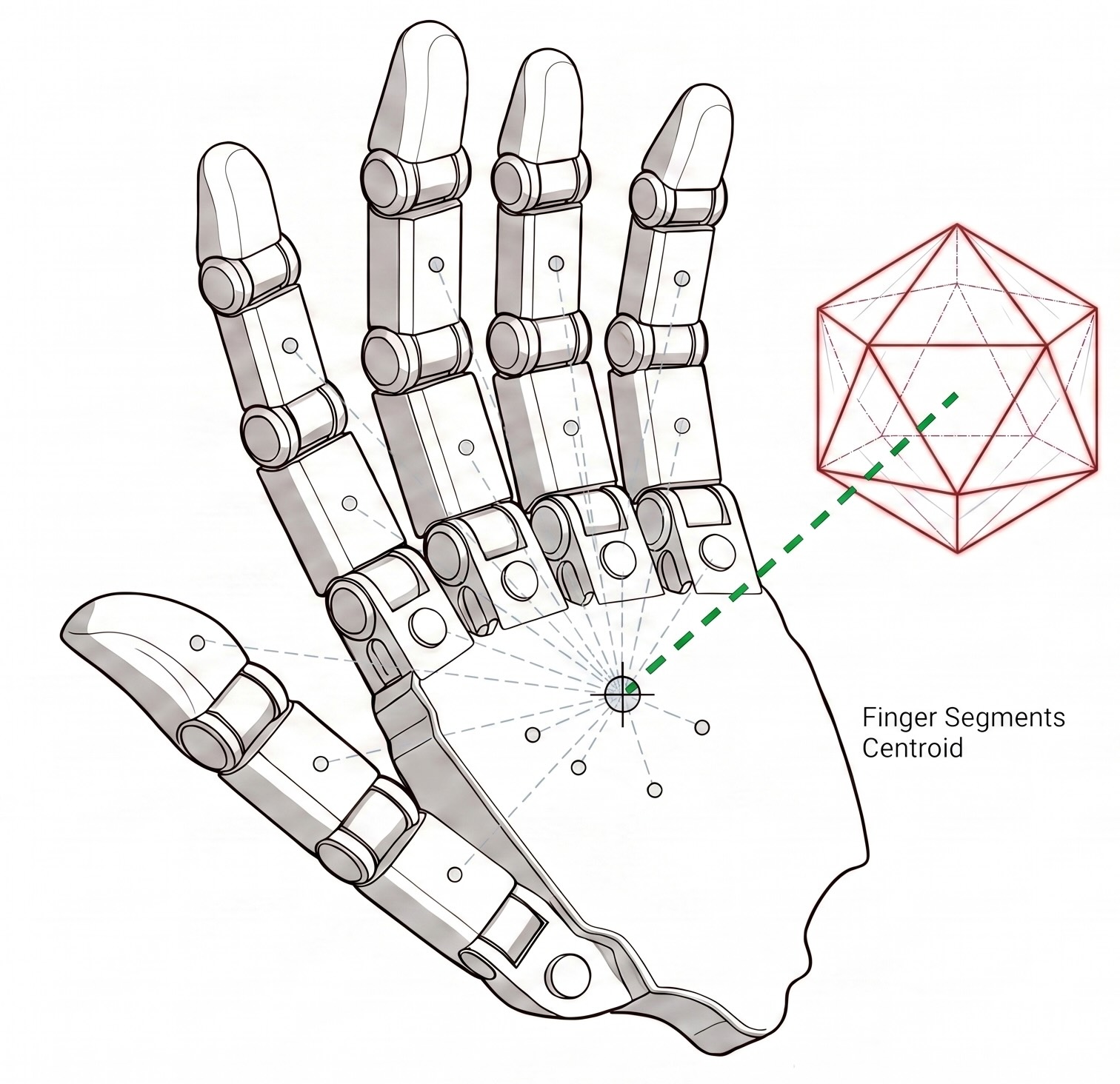}\label{fig:encapsulation}}
    \caption{Schematic illustrations of the hand-object interaction metrics. (a) Proximity of a fixed hand point to the object. (b) Alignment of finger surface normals with the object surface normal. (c) Proximity of finger segments to the object. (d) Proximity of sampled object points to the nearest finger. (e) Proximity of the finger-segment centroid to the object center.}
    \label{fig:interaction_scores}
\end{figure*}

We encourage spatial overlap by rewarding the proximity of the collective center of all finger segments to the center of the convex hull of the object, as shown in Fig.~\ref{fig:encapsulation}. We denote it as
$\exp{\left(-\alpha \bigl\|\mathbf{c}_{\text{obj}} - \mathbf{c}_{\text{fingers}}\bigr\|\right)}$.

If the center does not reside within the object, for instance in the case of the torus, a point inside the object is chosen. 

\subsection{Finger Joint Decoupling}
We apply an intra-episode curriculum to overcome the difficulty of high-dimensional control by pooling and gradually isolating the neural network policy output for the finger joints. This curriculum delays the increase in action-space dimensionality until after the policy has gained experience controlling in lower-dimensional space.

\begin{figure}[htbp]
    \centering
    \subfloat[Individual decoupling]{\includegraphics[width=0.45\linewidth]{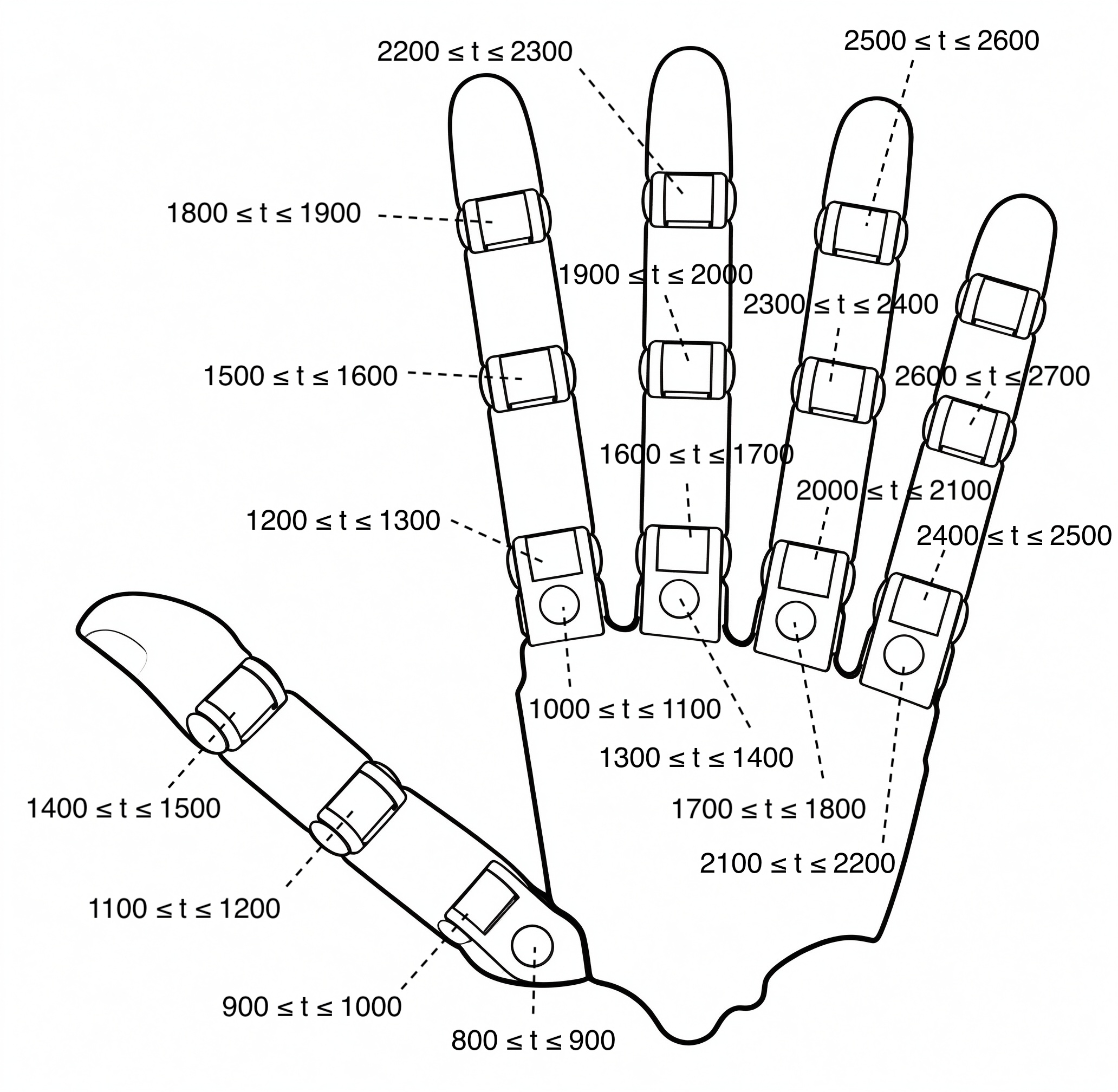}\label{decoupling_each_joint}}
    \hfil
    \subfloat[Group-based decoupling]{\includegraphics[width=0.45\linewidth]{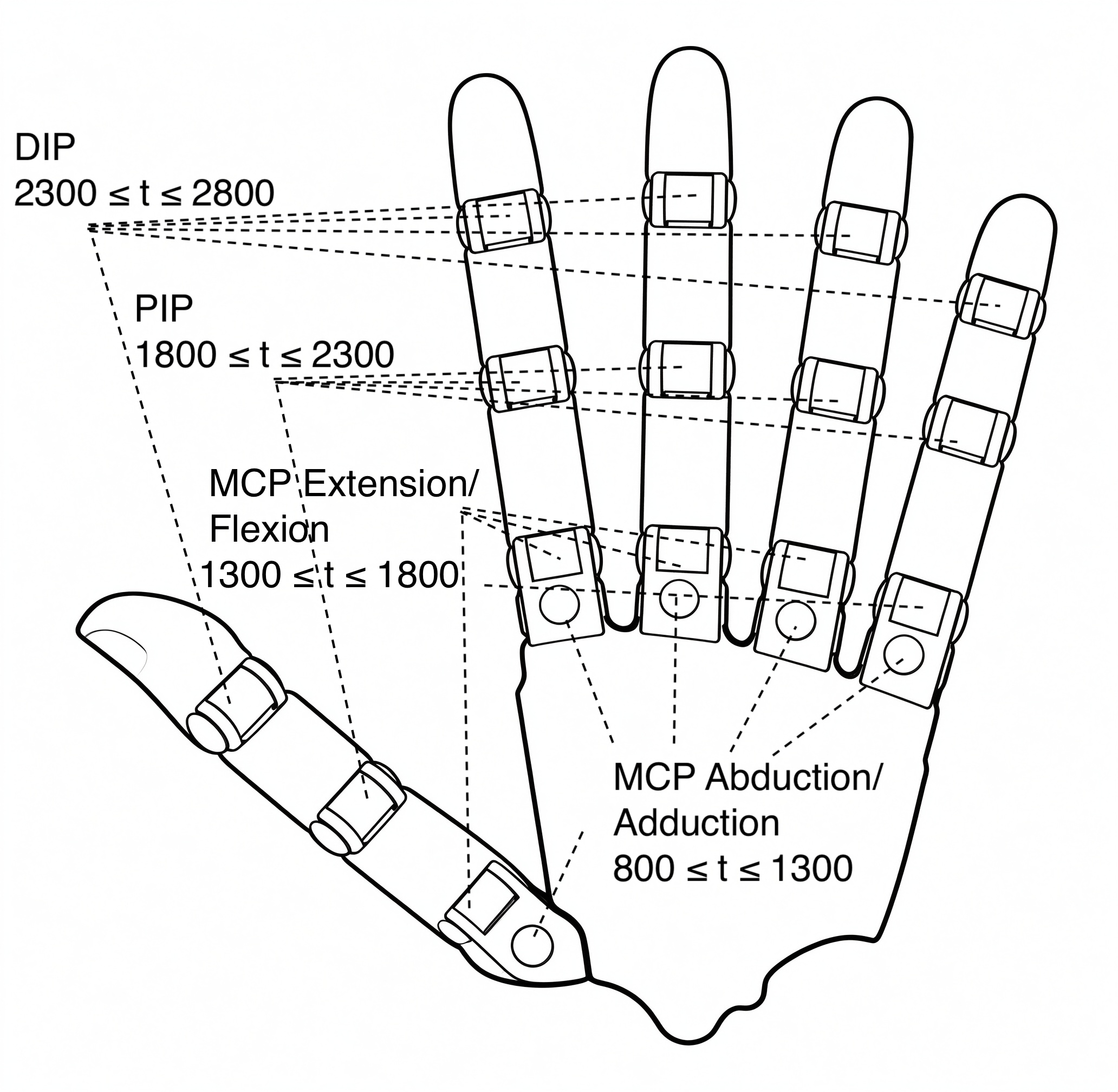}\label{decoupling_by_type}}
    \caption{Finger joint decoupling schedules. Each label $x \le t \le y$ indicates the time window during which the corresponding joint or joint group is linearly decoupled from the remaining averaged joints.}
    \label{fig:decoupling_schedules}
    \vspace{-5pt}
\end{figure}

At the beginning of each episode, all finger joints receive the averaged joint torque command. During each episode, finger joints are decoupled from the averaging function in a proximal-to-distal schedule. We examine two options:

\smallskip
\noindent {\bf Individual}: happens once every 100 time steps after $t=800$; the torque command of a finger joint is decoupled linearly from the remaining averaged joint torque commands according to the schedule in Fig.~\ref{decoupling_each_joint}.

\smallskip
\noindent
{\bf Group-based}: happens once every 500 time steps after $t=800$; the torque commands of the same joint types are decoupled together linearly from the remaining averaged joint torque commands according to Fig.~\ref{decoupling_by_type}.

After the schedule ends, all finger joints are controlled independently by the policy output.
\subsection{Achievement-triggered Rewards}
We train a whole-body reaching and grasping policy by breaking down the task into a few milestones inspired by developmental principles, such as rolling over, head-object alignment, hand-object proximity, encapsulation, etc. We define a reward function for each milestone and organize them as a directed acyclic graph, where the nodes are the rewards and the edges represent achievement conditions for ``unlocking'' the learning of subsequent ones.

We define two types of achievement scores: an enforced achievement score
\begin{equation*}
A_{i,j,t}^E = r_{i,t} - p_{i,j}
\end{equation*}
and a non-enforced achievement score
\begin{equation*}
A_{i,j,t}^N = \max_{t^*\le t} \bigl\{\,r_{i,t^*}\bigr\} - p_{i,j}
\end{equation*}
where $r_{i,t}$ is the instantaneous reward for milestone $i$ at time $t$, and $p_{i,j}$ is the passing score for milestone $i$ to activate milestone $j$. They differ in the reward value used for calculating the achievement score: the enforced one uses instantaneous reward, so $A_{i,j,t}^E$ can decrease once $r_{i,t}$ decreases. This is useful when modeling transitions where a prerequisite milestone such as pre-grasping needs to be maintained during a later milestone such as grasping. 

The non-enforced one uses max-over-all-time reward, so $A_{i,j,t}^N$ does not decrease even if $r_{i,t}$ decreases. This is useful when modeling transitions where a prior milestone such as hand-object proximity can be sacrificed once a later milestone such as finger-object proximity is being optimized.

\begin{figure*}[htbp]
    \centering
    \includegraphics[width=\textwidth]{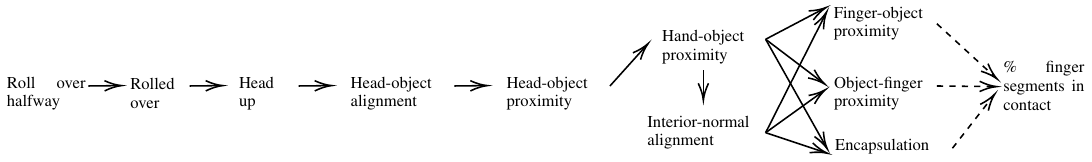}
    \caption{Achievement-triggered reward graph illustrating different transition mechanisms. Solid arrow lines indicate transitions using a non-enforced achievement score, whereas dashed arrow lines represent transitions with an enforced achievement score, which requires prerequisite milestones to remain activated.}
    \label{grasp_reward}
\end{figure*}

Generally, both types of achievement scores can be present in a reward graph. We use dashed arrow lines for transitions with enforced achievement scores and solid arrow lines for transitions using non-enforced achievement scores, as shown in Fig.~\ref{grasp_reward}. The reward function for each milestone is listed in Table~\ref{reward_functions} with its level and multiplier.

The total reward at time $t$ is the sum of all active milestone rewards, weighted by their achievement scores and a milestone level multiplier:
\begin{equation*}
r_t =\sum_{j}\pow(L_j, w)\sum_{i}\max\bigl(0,\,A_{i,j,t}\bigr)r_{j, t}
\end{equation*}
where $L_j$ is the maximum number of hops from the initial milestone node to milestone $j$ in the graph, and $w$ is a fixed positive exponent representing the scaling factor for the learner to gain higher rewards for more advanced milestones.

\begin{table*}[htbp]
\centering
\caption{Reward functions}
\label{reward_functions}
\renewcommand{\arraystretch}{1.5}
\begin{tabular}{l l c c c}
    \toprule
    \textbf{Milestone} & \textbf{Expression} & \textbf{Level} & \textbf{\makecell{Multiplier \\ ($w=2$)}} & \textbf{\makecell{Passing \\ score}} \\
    \midrule
Roll over halfway (lateral axis of root points upward) & $\hat{y}_{\textrm{root}, z}$ & 1 & 1 & 85\% \\
Rolled over (anterior axis of root points downward) & $- \hat{x}_{\textrm{root}, z}$ & 2 & 4 & 75\% \\
Head up (head's superior axis points upward) & $\hat{z}_{\textrm{head}, z}$ & 3 & 9 & 75\% \\
Head-object alignment (head's anterior axis points to object) & $\hat{x}_{\textrm{head}}\cdot\frac{o_\textrm{object}-o_\textrm{head}}{\bigl\| o_\textrm{object}-o_\textrm{head}\bigr\|}$ & 4 & 16 & 75\% \\
Head-object proximity (head at arm distance from object) & $\exp\bigl(-\alpha(\bigl\| o_\textrm{object}-o_\textrm{head}\bigr\| - \textrm{L}_\textrm{arm})^2\bigr)$ & 5 & 25 & 75\% \\
Hand-object proximity & $I(\textbf{h})$ & 6 & 36 & 85\% \\
Interior-normal alignment & $-\vec{\mathbf{n}_\mathrm{f}}\cdot\vec{\mathbf{n}_\mathrm{o}}$ & 7 & 49 & 50\% \\
Finger-object proximity & $\avg(I(\mathbf{f}))$ & 8 & 64 & 70\% \\
Object-finger proximity & $\avg_k\left(\exp{\bigl(-\alpha\;\min_{j}\bigl\|\mathbf{f}_j-\mathbf{\Omega}_k\bigr\|\bigr)}\right)$ & 8 & 64 & 50\% \\
Encapsulation & $\exp{\left(-\alpha \bigl\|\mathbf{c}_{\text{obj}} - \mathbf{c}_{\text{fingers}}\bigr\|\right)}$ & 8 & 64 & 40\% \\
\% of finger segments in contact & $\frac{\sum {\{\bigl\|F_\textrm{finger}\bigr\| > F_\textrm{threshold}\}}}{N_\textrm{fingers}}$ & 9 & 81 & 20\% \\
    \bottomrule
\end{tabular}
\end{table*}

\subsection{Policy Composition}
The policy composition module is implemented as a multi-policy RL player. When performing a composite task that requires multiple policies, such as reaching and grasping followed by standing and walking, we first construct a unified observation vector comprising the observation data needed for each subtask policy, as listed in Table~\ref{observation}. The policy composition module then extracts the independent observation vectors for each subtask policy from the unified observation vector via index selection.

At the beginning of the episode, the reaching and grasping policy commands all robot joints. After the object is grasped, indicated by passing the final milestone in the reward graph (\% of finger segments in contact), the getting-up policy takes over control of body joints excluding the fingers, to allow the robot to get up and walk while holding the object in hand.

\begin{table*}[htbp]
\centering
\caption{Observation vector components}
\label{observation}
\renewcommand{\arraystretch}{1.25}
\begin{tabular}{@{}l l l l l l@{}}
    \toprule
& & \multicolumn{4}{c}{\textbf{Dimension for policy and robot type}} \\
\cmidrule(r){3-6}
& & \textbf{Manipulation} & \multicolumn{3}{c}{\textbf{Locomotion}} \\
\cmidrule(lr){3-3} \cmidrule(lr){4-6}
    \textbf{Expression} & \textbf{Meaning} & \textbf{Full 72 DoF} & \textbf{Full 72 DoF} & \textbf{32 DoF, non-actuated fingers} & \textbf{32 DoF, no fingers} \\
    \midrule
$z_\mathrm{root}$ & height of robot root (pelvis) & 1 & 1 & 1 & 1 \\
$q_\mathrm{root}$ & orientation of robot root & 4 & 4 & 4 & 4 \\
$v_\mathrm{root}$ & velocity of robot root & 3 & 3 & 3 & 3 \\
$\omega_\mathrm{root}$ & angular velocity of robot root & 3 & 3 & 3 & 3 \\
$o_\mathrm{links}$ & origin of robot links and object & 83$\times$3 & 83$\times$3 & 83$\times$3 & 42$\times$3 \\
$q_\mathrm{links}$ & orientation of robot links and object & 83$\times$4 & 83$\times$4 & 83$\times$4 & 42$\times$4 \\
$v_\mathrm{links}$ & velocity of robot links and object & 83$\times$3 & 83$\times$3 & 83$\times$3 & 42$\times$3 \\
$F_\mathrm{fingers}$ & finger contact force & 40$\times$3 & 40$\times$3 & 40$\times$3 & None \\
$\mathrm{CHWSC}(6)$ & CHWSC of finger segments & 44$\times$2 & None & None & None \\
$q_\mathrm{joints}$ & joint angular position & 72 & 72 & 72 & 32 \\
$\omega_\mathrm{joints}$ & joint angular velocity & 72 & 72 & 72 & 32 \\
$F_\mathrm{joints}$ & joint force & 72 & 72 & 72 & 32 \\
$\tau_\mathrm{thigh}$ & force and torque at thigh & 12 & 12 & 12 & 12 \\
$\tau_\mathrm{upper\;arm}$ & force and torque at upper arm & 12 & 12 & 12 & 12 \\
$a_\mathrm{proc}$ & previous step's policy output & 72 & 72 & 32 & 32 \\
$a_\mathrm{proc}\tau_\mathrm{max}$ & previous step's joint torque command & 72 & 72 & 32 & 32 \\
$\sin\bigl(0.5\,\pi\,t \, f\bigr)$ & CPG signals & 16 & 16 & 16 & 16 \\
$\frac{t}{t_\mathrm{max}}$ & episode progress ratio & 1 & 1 & 1 & 1 \\
    \bottomrule
\end{tabular}
\end{table*}

\section{Experiments}
We implemented the method in a simulation environment and tested it with ablation studies. 

\subsection{Simulation Environment}
We run 4096 simulation environments in Isaac Gym~\cite{IsaacGym}. Each environment contains a humanoid robot and a cube. The robot model has 72 DoF in total, with each anthropomorphic hand having 20 DoF. The robots are initially lying on the ground in a supine position with random joint angles, while the cube is located on the ground in random directions. Each episode contains 3000 time steps. We apply domain randomization to robot link mass and friction, joint damping and stiffness, and gravity. Training hyper-parameters are listed in Table~\ref{hyper_parameters}. We use proximal policy optimization \cite{ppo} to train a neural network policy with 4 fully-connected layers of 1600, 800, 400, and 200 neurons, using ELU activation.
\begin{table}[!h]
\centering
\begin{threeparttable}
\renewcommand{\arraystretch}{1.1}
\caption{List of Hyper-parameters.}
\label{hyper_parameters}
\centering
\begin{tabular}{@{}l l | l l@{}}
\toprule
\textbf{Simulator} & & \textbf{Randomization} & \\
    \cmidrule(lr){1-2}\cmidrule(lr){3-4}
Timestep length & 0.005 s & Observations & $+\mathcal{N}(0, 0.001)$ \\
Position iterations & 4 & Actions & $+\mathcal{N}(0, 0.0075)$ \\
Velocity iterations & 0 & Gravity & $+\mathcal{N}(0, 0.175)$ \\
Contact offset & 0.01 m & Body mass & $\times \mathcal{U}\left [0.75, 1.25\right ]$ \\
Rest offset & 0.01 m & Friction & $\times \mathcal{U}\left [0.85, 1.15\right ]$ \\
Bounce threshold & 0.5 m/s & Damping \& stiffness & $\times \mathcal{U} \left [0.8, 1.2\right]$ \\
Max depenetration & 0.2 m/s & Joint limit & $+\mathcal{N}(0, 0.005)$ \\
\bottomrule
\vspace{-10pt}
\end{tabular}
\begin{tablenotes}
    \small
    \item $+\mathcal{N}(\textrm{mean}, \textrm{variance})$ stands for additive normal distribution.
    \item $\times \mathcal{U} \left [\textrm{lower}, \textrm{upper}\right]$ is scaling with uniform distribution.
\end{tablenotes}
\end{threeparttable}
\end{table}

We test the policy composition module using the grasping policy trained on the 72-DoF robot and getting-up policies trained on 3 variations of the robot listed in Table~\ref{composition_result}.

\subsection{Ablation}
To verify the effectiveness of having CHWSC in the observation vector, we experiment with cubic harmonics of degrees up to 6 and without CHWSC. We also compare the convoluted spatial representation with unconvoluted raw value by replacing CHWSC with finger-object proximity and direction vectors from the object to each finger.

To verify the importance of gradual finger joint decoupling, we run an experiment without the decoupling curriculum while still using CHWSC in the observation vector.

Finally, to show how these two methods work in synergy, we run a baseline experiment containing neither CHWSC nor finger joint decoupling, representing a naive RL environment.

\section{Results}
\begin{figure*}[htbp]
    \centering
    \subfloat[CHWSC(6), D(1)]{\includegraphics[width=0.32\linewidth]{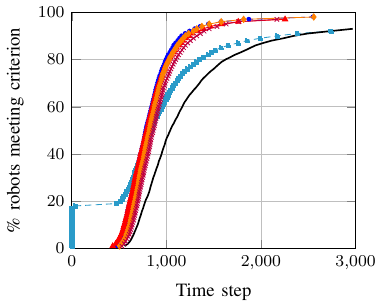}}\hfil
    \subfloat[CHWSC(5), D(1)]{\includegraphics[width=0.32\linewidth]{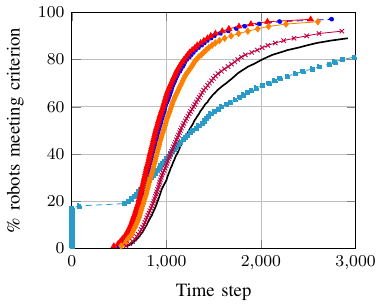}}\hfil
    \subfloat[CHWSC(4), D(1)]{\includegraphics[width=0.32\linewidth]{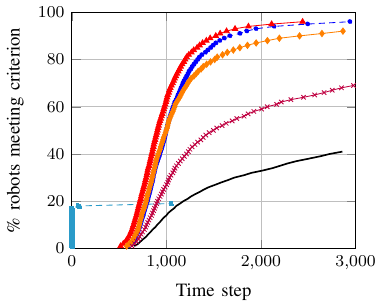}}

    \subfloat[CHWSC(3), D(1)]{\includegraphics[width=0.32\linewidth]{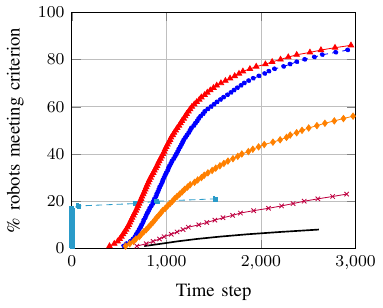}}\hfil
    \subfloat[CHWSC(2), D(1)]{\includegraphics[width=0.32\linewidth]{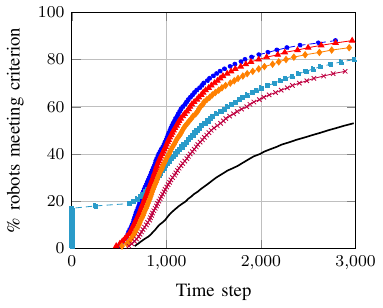}}\hfil
    \subfloat[CHWSC(1), D(1)]{\includegraphics[width=0.32\linewidth]{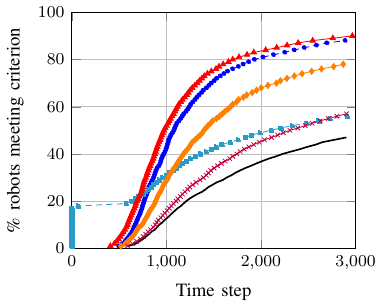}}

    \subfloat[CHWSC(0), D(1)]{\includegraphics[width=0.32\linewidth]{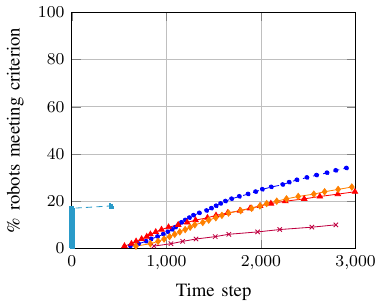}}\hfil
    \subfloat[CHWSC(6), D(5)]{\includegraphics[width=0.32\linewidth]{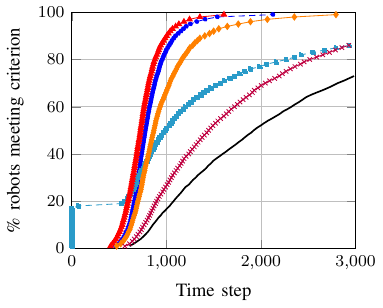}}\hfil
    \subfloat[No CHWSC, D(1)]{\includegraphics[width=0.32\linewidth]{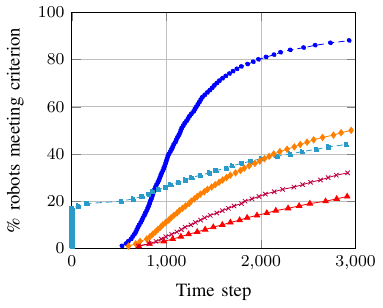}}

    \subfloat[Unconvoluted, D(1)]{\includegraphics[width=0.32\linewidth]{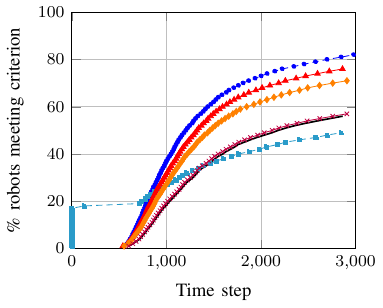}}\hfil
    \subfloat[CHWSC(6), No D]{\includegraphics[width=0.32\linewidth]{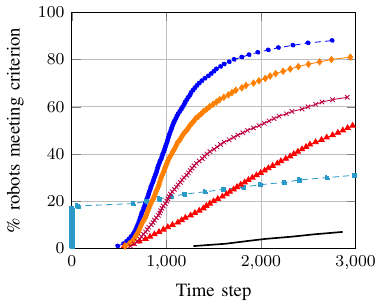}}\hfil
    \subfloat[Neither]{\includegraphics[width=0.32\linewidth]{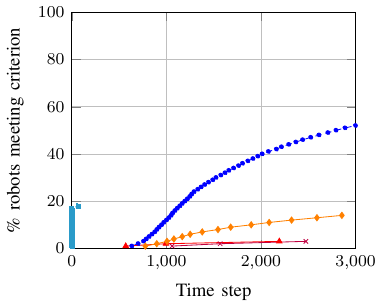}}

    \includegraphics[width=\linewidth]{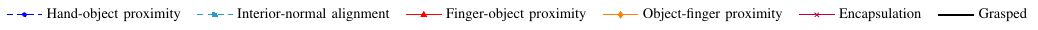}%
    \caption{Percentage of robots that have passed each hand-object interaction score. Each sub-figure shows an episode of the policy trained from each experiment. Dashed black line represents the grasp success rate (finger-object proximity, object-finger proximity, and encapsulation scores passed).}
    \label{fig:all_experiments}
\end{figure*}

Our methods successfully trained a humanoid robot to roll over from a supine pose to a prone pose, crawl toward an object, reach out its arms, and grasp the objects. The policy composition module then commands the robot to stand up and walk while holding the object. Table~\ref{tab:success_rate_learning_speed} shows the success rates of using different degrees of cubic harmonics versus ablation experiments. A successful grasp is defined as passing the finger-object proximity, object-finger proximity, and encapsulation scores. Fig.~\ref{fig:all_experiments} is a detailed breakdown showing the percentage of robots that have passed each hand-object interaction score during policy execution.

The results show that both CHWSC and finger joint decoupling are critical for the policy to learn grasping. As the maximum degree of cubic harmonics decreases, the success rate drops from 93\% to 0\%, suggesting that higher-order spatial convolution benefits learning by providing a more informative representation of the hand-object interaction relation. Without CHWSC (by removing it or using unconvoluted distance and direction of the fingers with respect to the object), the observation vector cannot provide enough spatial information for the policy to learn grasping effectively.

Without joint decoupling, only 7\% of robots succeeded in grasping, suggesting that the RL algorithm suffers from a high action space dimension; the policy was unable to learn to control all 72 DoF at once. With finer joint decoupling, the policy only needs to learn 32 DoF at the beginning and gradually learns to control a 72-DoF action space later in the training stage. Decoupling the joints in groups performs better for proximity but worse for encapsulation and overall success rate than decoupling individually.

If neither CHWSC nor joint decoupling is used, the grasp success rate was 0\%, and the policy stopped learning at only 7,407 epochs, suggesting that the RL algorithm was unable to learn any rewarding actions due to the missing spatial information and the high dimensionality of the action space.

Table~\ref{zero_shot_result} shows the zero-shot grasp success rates of unseen objects. Although the policy was only trained using a cube, its generalizability shows the benefit of learning to interact with a distance field rather than explicit object geometry.

\begin{table}[htbp]
\centering
\caption{Zero-shot grasp success rates for unseen objects.}
\label{zero_shot_result}
\begin{tabular}{@{}lr|lr@{}}
\toprule
\textbf{Object} & \textbf{Success rate} &\textbf{Object} & \textbf{Success rate} \\
\midrule
Sphere & 98\% & Tetrahedron & 94\% \\
Cuboid & 94\% & Octahedron & 93\% \\
Cylinder & 93\% & Icosahedron & 94\% \\
\bottomrule
\end{tabular}
\end{table}
\begin{table}[htbp]
\centering
\caption{Success rates of standing up and walking with object being grasped with the standing-up policy trained on different robot models.}
\label{composition_result}
\begin{tabular}{@{}lc@{}}
\toprule
\textbf{Robot model for stand-up policy training} & \textbf{Success rate} \\
\midrule
Same robot, 72 DoF & 100\% \\
Same robot, non-actuated fingers, 32 DoF & 96\% \\
Robot without fingers, 32 DoF & 31\% \\
& (76\% got up but 45\% fell) \\
\bottomrule
\vspace{-5pt}
\end{tabular}
\end{table}

Table~\ref{composition_result} shows the success rates of standing up and walking when composing the grasping policy with the standing-up policy. Interestingly, although all three standing-up policies had a 100\% success rate on the embodiment they were trained on, only the policies trained on the full robot model with fingers were able to stand up when composed with the grasping policy trained on the full 72-DoF robot.

The comparative results imply that whether observation space and robot mechanism match or not determines cross-embodiment transferability between different policies. To combine standing-up policy with reaching out and grasping policies, the standing-up policy learned on a robot without fingers could not work. The composition only works if the standing-up policy learned also on the robot with fingers.

\begin{table*}[htbp]
\centering
\caption{Grasp success rate and learning speed across ablation conditions.}
\label{tab:success_rate_learning_speed}
\renewcommand{\arraystretch}{1.1}
\begin{tabular}{@{}lcccccccc@{}}
\toprule
& & & \multicolumn{3}{c}{\textbf{\% robots meeting criterion}} & \\
\cmidrule(r){4-6}
\textbf{Experiment ID} & \makecell{Cubic harmonics \\ degree} & Joint decoupling & \makecell{Finger-object \\ proximity} & \makecell{Object-finger \\ proximity} & Encapsulation & \textbf{\makecell{Success \\ rate}} & \textbf{\makecell{Total \\ reward}} & \textbf{\makecell{Training \\ Epochs}} \\
\midrule
CHWSC(6), D(1) & $0\leq\ell\leq 6$ & Individual & \textbf{97\%} & \textbf{98\%} & \textbf{97\%} & \textbf{93\%} & \textbf{414\,376} & \textbf{23\,907} \\
CHWSC(5), D(1) & $0\leq\ell\leq 5$ & Individual & 97\% & 96\% & 92\% & 89\% & 421\,887 & 32\,813 \\
CHWSC(4), D(1) & $0\leq\ell\leq 4$ & Individual & 96\% & 92\% & 69\% & 41\% & 278\,800 & 33\,094 \\
CHWSC(3), D(1) & $0\leq\ell\leq 3$ & Individual & 86\% & 56\% & 23\% & 8\% & 221\,712 & 11\,344 \\
CHWSC(2), D(1) & $0\leq\ell\leq 2$ & Individual & 88\% & 85\% & 75\% & 53\% & 224\,928 & 33\,750 \\
CHWSC(1), D(1) & $0\leq\ell\leq 1$ & Individual & 90\% & 78\% & 57\% & 47\% & 233\,171 & 24\,000 \\
CHWSC(0), D(1) & $\ell=0$ & Individual & 24\% & 26\% & 10\% & 0\% & 116\,098 & 8\,344 \\
\midrule
CHWSC(6), D(5) & $0\leq\ell\leq 6$ & Group-based & 99\% & 99\% & 86\% & 73\% & 434\,384 & 35\,719 \\
\midrule
No CHWSC, D(1) & & Individual & 22\% & 50\% & 32\% & 0\% & 240\,376 & 37\,782 \\
\midrule
Unconvoluted, D(1) & & Individual & 76\% & 71\% & 57\% & 56\% & 211\,278 & 36\,563 \\
\midrule
CHWSC(6), No D & $0\leq\ell\leq 6$ & & 52\% & 81\% & 64\% & 7\% & 259\,932 & 39\,750 \\
\midrule
Neither & & & 3\% & 14\% & 3\% & 0\% & 161\,759 & 7\,407 \\
\bottomrule
\vspace{-10pt}
\end{tabular}
\end{table*}

Qualitative outcomes and the corresponding video are provided in the supplementary material.

\section{Conclusions}

\noindent {\bf Summary}:
This work introduced an autonomous developmental reinforcement learning framework that enables a humanoid robot to learn a reusable whole-body reaching and grasping policy efficiently without requiring external datasets or pre-trained models. It further demonstrated a policy composition method to combine separately trained skill policies for more sophisticated composite robot behaviors. 

The introduced CHWSC provides a concise and informative spatial representation of hand-object interaction relations, allowing the robot to learn to grasp various types of objects and generalize to unseen objects.

Gradual finger joint decoupling and achievement-triggered rewards with enforcement provide a curriculum to the learning algorithm, enabling the robot to learn motor skills progressively and refine its dexterity.

Policy composition confirms that diverse motor policies trained on the same robot model can be seamlessly integrated at the inference time, implying that it is feasible to construct complex skills by combining multiple simpler motor policies.

\smallskip
\noindent {\bf Some future extensions}:
This work does not explicitly use object shape in training the robot reaching and grasping policies, but the object shape information was used in some computations. In real-world deployment, object shapes can be approximated from multiple visual perceptions~\cite{KinectFusion}. To further enhance generalizability, more types of objects can be included as a training curriculum. 

The current hand-object interaction scores used are irrespective of whether the left or right hand should be used. If a task requires a specific hand or bimanual manipulation, the interaction scores can be defined separately for each hand.

The policy composition module currently does not address picking up the object if it is dropped when the getting-up policy has already taken over control. Large language reasoning models may be utilized to select the individual motor policies under appropriate conditions.

\bibliography{ref-full}

\end{document}